%% file: paper1133.tex
\documentclass[runningheads]{llncs}
\usepackage{graphicx}
\usepackage{caption}
\usepackage{subfigure}
\usepackage{color}
\usepackage{ulem}
\usepackage{textcomp}
\usepackage{tipa}
\usepackage{amssymb}
\usepackage{dsfont}
\usepackage{subfigure}
\usepackage{mathrsfs}
\usepackage{bm}
\usepackage{multirow}
\usepackage{amsmath}
\usepackage{cite}
\usepackage[colorlinks,linkcolor=blue,anchorcolor=blue,citecolor=blue]{hyperref}
\newcommand{\samethanks}[1][\value{footnote}]{\footnotemark[#1]}

\begin{document}
%
% \title{Contribution Title\thanks{Supported by organization x.}}
\title{Shape-aware Semi-supervised 3D Semantic Segmentation for Medical Images}

% \author{Anonymous\inst{}}
%
\author{Shuailin Li \thanks{Both authors contributed equally to the work. This work was supported by Shanghai NSF Grant (No. 18ZR1425100).} \inst{1} \and 
Chuyu Zhang \samethanks \inst{1} \and
Xuming He\inst{1,2}}
% index{Li, Shuailin}, index{Zhang, Chuyu} , index{He, Xuming} 

%
% \authorrunning{Anonymous}
% \institute{Anonymous Organization \\\email{***@********.***}}
\institute{
    ShanghaiTech University, Shanghai, China \\ \and
    Shanghai Engineering Research Center of Intelligent Vision and Imaging \\
    \email{\{lishl, hexm\}@shanghaitech.edu.cn,
    zcy\_ai@whu.edu.cn}
    }

%\institute{ShanghaiTech University, SHANGHAI CHINA\\
%\email{\{Anonymous\}@shanghaitech.edu.cn}}
\maketitle              % typeset the header of the contribution

\input{section/abstract.tex}
\input{section/intro.tex}
\input{section/method.tex}
\input{section/experiment.tex}

\input{section/conclusion.tex}
\bibliographystyle{splncs04}
\bibliography{paper1133}

%     \bibitem{ref_article1}
%     Author, F.: Article title. Journal \textbf{2}(5), 99--110 (2016)

%     \bibitem{ref_lncs1}
%     Author, F., Author, S.: Title of a proceedings paper. In: Editor,
%     F., Editor, S. (eds.) CONFERENCE 2016, LNCS, vol. 9999, pp. 1--13.
%     Springer, Heidelberg (2016). \doi{10.10007/1234567890}

%     \bibitem{ref_book1}
%     Author, F., Author, S., Author, T.: Book title. 2nd edn. Publisher,
%     Location (1999)

%     \bibitem{ref_proc1}
%     Author, A.-B.: Contribution title. In: 9th International Proceedings
%     on Proceedings, pp. 1--2. Publisher, Location (2010)

%     \bibitem{ref_url1}
%     LNCS Homepage, \url{http://www.springer.com/lncs}. Last accessed 4
%     Oct 2017
% \end{thebibliography}
\end{document}

%% file: section/abstract.tex
\begin{abstract}
    Semi-supervised learning has attracted much attention in medical image segmentation due to challenges in acquiring pixel-wise image annotations, which is a crucial step for building high-performance deep learning methods. Most existing semi-supervised segmentation approaches either tend to neglect geometric constraint in object segments, leading to incomplete object coverage, or impose strong shape prior that requires extra alignment. 
    In this work, we propose a novel shape-aware semi-supervised segmentation strategy to leverage abundant unlabeled data and to enforce a geometric shape constraint on the segmentation output. To achieve this, we develop a multi-task deep network that jointly predicts semantic segmentation and signed distance map (SDM) of object surfaces. During training, we introduce an adversarial loss between the predicted SDMs of labeled and unlabeled data so that our network is able to capture shape-aware features more effectively. Experiments on the Atrial Segmentation Challenge dataset show that our method outperforms current state-of-the-art approaches with improved shape estimation, which validates its efficacy. Code is available at \url{https://github.com/kleinzcy/SASSnet}.

    \keywords{Geometric constraints  \and Semantic segmentation \and Semi-supevised learning.}
\end{abstract}

%% file: section/intro.tex
\section{Introduction}

%background: segmentation - deep network - labeling cost
Semantic object segmentation is a fundamental task in medical image analysis and has been widely used in automatic delineation of regions of interest in 3D medical images, such as cells, tissues or organs. Recently, tremendous progress has been made in medical semantic segmentation~\cite{taghanaki2019deep} thanks to modern deep convolutional networks, which achieve state-of-the-art performances in many real-world tasks. However, training deep neural networks often requires a large amount of annotated data, which is particularly expensive in medical segmentation problems. 
In order to reduce labeling cost, a promising approach is to adopt a semi-supervised learning~\cite{bai2017semi,baur2017semi} framework that typically utilizes a small labeled dataset and many unlabeled images for effective model training.    

%literature survey: what have been done in semi
Recent efforts in semi-supervised segmentation have been focused on incorporating unlabeled data into convolutional network training, which can be largely categorized into two groups. The first group of those methods mainly consider the generic setting of semi-supervised segmentation~\cite{zhang2017deep,hung2019adversarial,nie2018asdnet,zheng2019semi,laine2016temporal, tarvainen2017mean, yu2019uncertainty,bortsova2019semi,li2018semi}. Most of them adopt adversarial learning or consistency loss as regularization in order to leverage unlabeled data for model learning. The adversarial learning methods~\cite{zhang2017deep, hung2019adversarial,nie2018asdnet,zheng2019semi} enforces the distributions of segmentation of unlabeled and labeled images to be close while the consistency loss approaches~\cite{laine2016temporal,tarvainen2017mean,yu2019uncertainty,bortsova2019semi,li2018semi} utilize a teacher-student network design and require their outputs being consistent under random perturbation or transformation of input images. To cope with difficult regions, Nie et al.~\cite{nie2018asdnet} utilize adversarial learning to select regions of unlabeled images with high confidence to train the segmentation network. 
Yu et al.~\cite{yu2019uncertainty} introduce an uncertainty map based on the mean-teacher framework~\cite{tarvainen2017mean} to guide student network learning. Despite their promising results, those methods lack explicit modeling of the geometric prior of semantic objects, often leading to poor object coverage and/or boundary prediction.  
       
The second group of semi-supervised methods attempt to address the above drawback by incorporating a strong anatomical prior on the object of interest in their model learning~\cite{zheng2019semi, he2019dpa}. For instance, Zheng et al.~\cite{zheng2019semi} introduce the Deep Atlas Prior (DAP) model that encodes a probabilistic shape prior in its loss design. He et al.~\cite{he2019dpa} propose an auto-encoder to learn priori anatomical features on unlabeled dataset. However, such prior typically assumes properly aligned input images, which is difficult to achieve in practice for objects with large variation in pose or shape. 

%Our main idea
In this work, we propose a novel shape-aware semi-supervised segmentation strategy to address the aforementioned limitations. Our main idea is to incorporate a more flexible geometric representation in the network so that we are able to enforce a global shape constraint on the segmentation output, and meanwhile to handle objects with varying poses or shapes. Such a ``shape-aware" representation enables us to capture the global shape of each object class more effectively. Moreover, by exploiting consistency of the geometric representations between labeled and unlabeled images, we aim to design a simple and yet effective semi-supervised learning strategy for deep segmentation networks.

%Here we use "shape-aware" to emphasize that our method is able to capture the global shape of 3D object category more effectively.

%Traditional segmentation approaches adopt post processing for smooth and continuous shape.To incorporate geometric regularizations into neural networks, recently some approaches are proposed. Based on distance transform, Kervadec et al. \cite{kervadec2018boundary} reformulates a boundary loss term as a regional integral, and Karimi et al. \cite{D_HDloss_2020} designs a hausdorff loss, which provide complementary information to regional losses.Besides simply adding an extra boundary-aware loss, to capture the overall shape of the target organs, 
%Xue et al. \cite{xue2019shape} converts the segmentation task into predicting an signed distance map, which encodes richer structure information. 

%Detail pipeline
To achieve this, we develop a multi-task deep network that jointly predicts semantic segmentation and signed distance map (SDM)~\cite{perera2015motion, dangi2019distance, park2019deepsdf, xue2019shape} with a shared backbone network module. The SDM assigns each pixel a value indicating its signed distance to the nearest boundary of target object, which provides a shape-aware representation that encodes richer features of object shape and surface. To utilize the unlabeled data, we then introduce an adversarial loss between the predicted SDMs of labeled and unlabeled data for semi-supervised learning. This allows the model to learn shape-aware features more effectively by enforcing similar distance map distributions on the entire dataset. In addition, the SDM naturally imposes more weights on the interior region of each semantic class, which can be viewed as a proxy of confidence measure. In essence, we introduce an implicit shape prior and its regularization based on an adversarial loss for semi-supervised volumetric segmentation.  

We evaluate our approach on the Atrial Segmentation Challenge dataset with extensive comparisons to prior arts. The results demonstrate that our segmentation network outperforms the state-of-the-art methods and generates object segmentation with high-quality global shapes.

%In a word, we introduce implicit shape prior and its regularization with adversarial loss in the task of semi-supervised volumetric segmentation. Such a strategy, to our best knowledge, has not been explored before.

Our main contributions are three-folds: (1) We propose a novel shape-aware semi-supervised segmentation approach by enforcing geometric constraints on labeled and unlabeled data. (2) We develop a multi-task loss on segmentation and SDM predictions, and impose global consistency in object shapes through adversarial learning. (3) Our method achieves strong performance on the Atrial Segmentation Challenge dataset with only a small number of labeled data.

%% file: section/method.tex
\section{Method}
    % figure 1: model
    \begin{figure}[t!]
    	\centering 
        \includegraphics[width=0.9\textwidth]{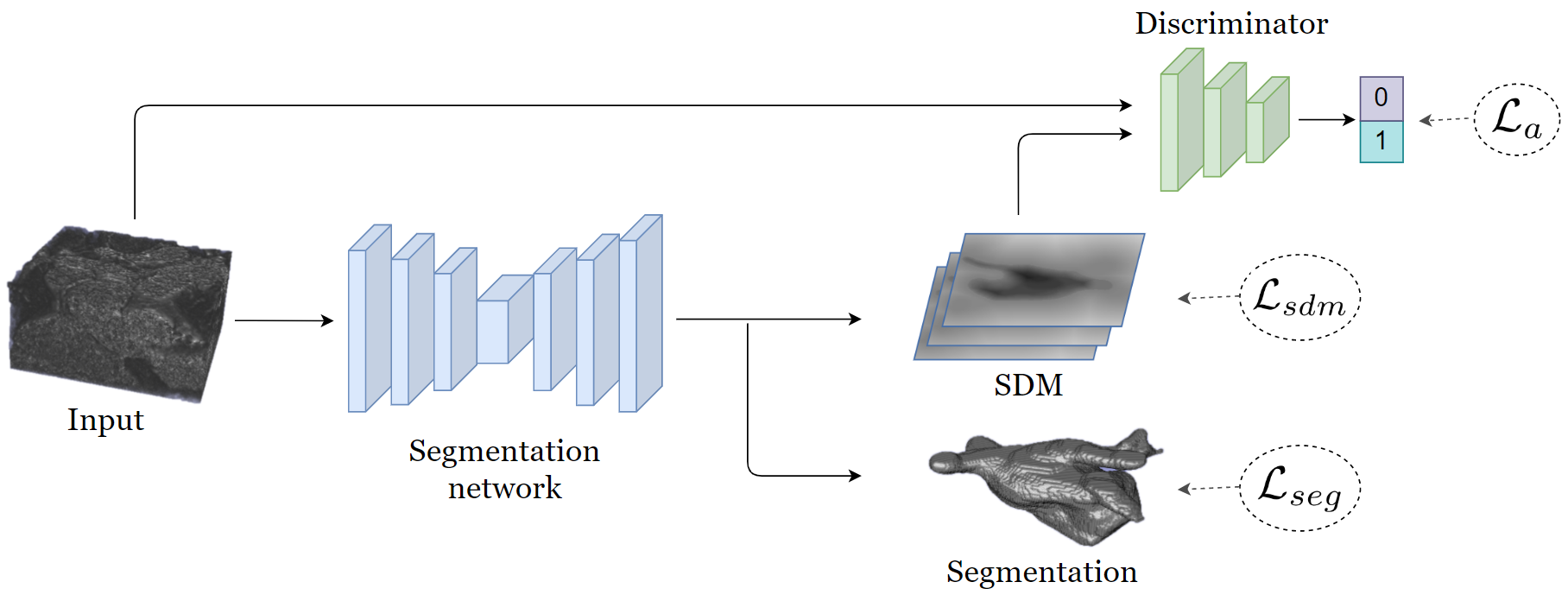}
        \caption{Overview of our method. Our network takes as input a 3D volume, and predicts a 3D SDM and a segmentation map. Our learning loss consists of a multi-task supervised term and an adversarial loss on the SDM predictions. 
        } 
        \label{fig:model}
    \end{figure}
    
\subsection{Overview}

We aim to build a deep neural network for medical image segmentation in a semi-supervised setting in order to reduce annotation cost. Due to lack of annotated images, our key challenge is to regularize the network learning effectively from a set of unlabeled ones. 
In this paper, we tackle this problem by utilizing the regularity in geometric shapes of the target object class, which provides an effective constraint for both segment prediction and network learning.  

Specifically, we propose to incorporate a shape-aware representation of object segments into the deep network prediction. In particular, we develop a multi-task segmentation network that takes a 3D image as input and jointly predicts a segmentation map and a SDM of object segmentation. Based on this SDM representation, we then design a semi-supervised learning loss for training the segmentation network. Our loss mainly consists of two components, one for the network predictions on the labeled set while the other enforcing consistency between the SDM predictions on the labeled and unlabeled set. To achieve effective consistency constraint, we adopt an adversarial loss that encourages the segmentation network to produce segment predictions with similar distributions on both datasets. Figure \ref{fig:model} illustrates the overall pipeline of our semi-supervised segmentation network. Below we will introduce the detailed model design in Section~\ref{sec:model}, followed by the learning loss and network training in Section~\ref{sec:loss}. 

%When this model is well trained, we not only improve the performance by utilizing labeled and unlabeled data, but also obtain segmentation maps with high-quality shape in 3D space.

\subsection{Segmentation Network}\label{sec:model}
% SN: function, io
% SN: seg calculation, sdm calculation (add reference)
% SN: implementation
%\subsubsection{Segmentation Network}

In order to encode geometric shape of a target semantic class, we propose a multi-task segmentation network that jointly predicts a 3D object mask and its SDM for the input 3D volume. Our network has a V-Net \cite{milletari2016v} structure that consists of an encoder module and a decoder module with two output branches, one for the segmentation map and the other for the SDM. For notation clarity, we mainly focus on the single-class setting below\footnote{It is straightforward to generalize our formulation to the multi-class setting by treating each semantic class separately for SDMs.}.

Specifically, we employ a V-Net backbone as in~\cite{yu2019uncertainty}, and then add a light-weighted SDM head in parallel with the original segmentation head. Our SDM head is composed by a 3D convolution block followed by the $tanh$ activation.
Given an input image $\mathbf{X}\in \mathbb{R}^{H\times W\times D}$, the segmentation head generates a confidence score map $\mathbf{M}\in [0,1]^{H\times W\times D}$ and the SDM head predicts a SDM $\mathbf{S}\in [-1,1]^{H\times W\times D}$ as follows:
\begin{align}
\mathbf{M} = f_\text{seg}(\mathbf{X}; \theta), \quad \quad  \mathbf{S} = f_\text{sdm}(\mathbf{X}; \theta)
\end{align}
where $\theta$ are the parameters of our segmentation network, and each element of $\mathbf{S}$ indicates the signed distance of a corresponding voxel to its closest surface point after normalization~\cite{xue2019shape}. %We generate the final semantic object masks based on the segmentation output $\mathbf{M}$ and use the SDM output $\mathbf{S}$ as our shape-aware representation for efficient semi-supervised learning.     

% Shuailin: implementation
%The segmentation network tries to generate results of unlabeled data look like labeled data. That is to say, we apply geometric constraint to unlabeled data indrectly. 
%Without groundtruth, we cannot apply geometric constraints for unlabeled data directly. To regularize the unlabeled data, we introduce a discriminator network that enforces the segmentation predictions of unlabeled data look like labeled data. Basically, adversarial learning approach is applied here. Compared to mean-teacher \cite{tarvainen2017mean} that forces pixel level consistency under perturbation, our discriminator enforces global consistency between labeled and unlabeled data.
%The input volume and SDM output are taken as inputs, and classified into binary labels through the discriminator. 
%In label space, 1 means that inputs are in labeled set and 0 otherwise. We design our discriminator based on \cite{radford2015unsupervised}, which consists of 5 convolution layers with 4×4 kernel and \{64, 128, 256, 512, 1\} channels in the stride of 2, and each convolution layer is followed by a Leaky-ReLU parameterized by 0.2 except the last layer. To fully utilize information from the input volume and SDM, we first encode the two into the same feature space, then fuse them for subsequent layers. In practice, we replace the first convolution layer by two separate convolution layers, which generate feature maps of the volume and SDM respectively. Then we add two feature maps and feed them into subsequent convolution layers.

\subsection{Shape-aware Semi-supervised Learning}\label{sec:loss}

% Chuyu: loss on labeled data: L_seg, L_sdm, L_D
%        loss on unlabeled data: L_adv, l_D
% 先总写 L = L_labled + L_unlabeled, 再分写 L_label=..., L_unlabeled=..
% 不需要引入 S()来表示SDM了，这个带来了混淆。

We now introduce our semi-supervised learning strategy for the segmentation network. While prior methods typically rely on the segmentation output $\mathbf{M}$, we instead utilize the shape-aware representation $\mathbf{S}$ to regularize the network training. To this end, we develop a multi-task loss consisting of a supervised loss $\mathcal{L}_{s}$  on the labeled set and an adversarial loss $\mathcal{L}_{a}$ on the entire set to enforce consistency of the model predictions.  

Formally, we assume a standard semi-supervised learning setting, in which the training set contains $N$ labeled data and $M$ unlabeled data, where $N\ll M$. We denote the labeled set as $\mathcal{D}^l=\{\mathbf{X}_n, \mathbf{Y}_n, \mathbf{Z}_n\}^N_{n=1}$ and unlabeled set as $\mathcal{D}^u=\{\mathbf{X}_m\}^{N+M}_{m=N+1}$, where $\mathbf{X}_n \in \mathbb{R} ^{H \times W \times D}$ are the input volumes,  $\mathbf{Y}_n \in \{0, 1\}^{H \times W \times D} $ are the segmentation annotations and $\mathbf{Z}_n \in \mathbb{R} ^{H \times W \times D}$  are the groundtruth SDMs derived from $\mathbf{Y}_n$. Below we first describe the supervised loss on $\mathcal{D}^l$ followed by the adversarial loss that utilizes the unlabeled set $\mathcal{D}^u$. 

\subsubsection{Supervised Loss $\mathcal{L}_{s}$}

On the labeled set, we employ a dice loss $l_{dice}$ and a mean square loss $l_{mse}$ for the segmentation and SDM output of the multi-task segmentation network, respectively:
\begin{align}
\mathcal{L}_{s}(\theta) =& \mathcal{L}_{seg} + \alpha \mathcal{L}_{sdm} \label{eq:supervised_loss} \\
\mathcal{L}_{seg} = \frac{1}{N}\sum_{i=1}^{N} l_{dice}(f_{seg}(\mathbf{X}_{i} ; \theta),& \mathbf{Y}_{i});\quad  %\label{eq:seg_loss} \\
\mathcal{L}_{sdm} = \frac{1}{N}\sum_{i=1}^{N} l_{mse}(f_{sdm}(\mathbf{X}_{i} ; \theta), \mathbf{Z}_{i}) %\label{eq:sdm_loss}
\end{align}
where $\mathcal{L}_{seg}$ denotes the segmentation loss and $\mathcal{L}_{sdm}$ is the SDM loss, and $\alpha$ is a weighting coefficient balancing two loss terms. 

\subsubsection{Adversarial Loss $\mathcal{L}_{a}$}

To regularize the model learning with the unlabeled data, we introduce an adversarial loss that enforces the consistency of SDM predictions on the labeled and unlabeled set. To this end, we propose a discriminator network to tell apart the predicted SDMs from the labeled set, which should be high-quality due to the supervision, and the ones from the unlabeled set. Minimizing the adversarial loss induced by this discriminator enables us to learn effective shape-aware features that generalizes well to the unlabeled dataset.   

Specifically, we adopt a similar discriminator network $D$ as~\cite{radford2015unsupervised}, which consists of 5 convolution layers followed by an MLP. The network takes a SDM and input volume as input, fuses them through convolution layers, and predicts its class probability of being labeled data. Given the discriminator $D$, we denote its parameter as $\zeta$ and define the adversarial loss as follows,
\begin{align}
\mathcal{L}_{a}(\theta,\zeta) = \frac{1}{N}\sum _{n=1}^{N} \log D(\mathbf{X}_n, \mathbf{S}_n; \zeta) +
\frac{1}{M}\sum_{m=N+1}^{N+M}\log\big(1-D(\mathbf{X}_m,\mathbf{S}_m;\zeta)\big) \label{eq:adv_loss}
\end{align}
where $\mathbf{S}_n=f_{sdm}(\mathbf{X}_n; \theta)$ and $\mathbf{S}_m=f_{sdm}(\mathbf{X}_m; \theta)$ are the predicted SDMs.

\subsubsection{Overall Training Pipeline}

Our overall training objective $\mathcal{V}(\theta, \zeta)$ combines the supervised and the adversarial loss defined above and the learning task can be written as, 
\begin{align}
\min_{\theta}\max_{\zeta}\mathcal{V}(\theta, \zeta) = \mathcal{L}_{s}(\theta) + \beta \mathcal{L}_{a}(\theta,\zeta)
\label{eq:total_loss}
\end{align}
where $\beta$ is a weight coefficient that balances two loss terms. We adopt a standard alternating procedure to train the entire network, which includes the following two subproblems. 

%\paragraph{Training segmentation network.}  
Given a fixed discriminator $D(\cdot;\zeta)$, we minimize the overall loss w.r.t the segmentation network parameter $\theta$. To speed up model learning, we simplify the loss in two steps: Firstly, we ignore the first loss term in Eqn~\eqref{eq:adv_loss} due to high-quality SDM predictions on the labeled set, i.e., $\mathbf{S}_n\approx \mathbf{Z}_n$, and additionally, we adopt a similar surrogate loss for the generator as in~\cite{Ian2014gan}. Hence the learning problem for the segmentation network can be written as,
  \begin{align}
        \min_\theta \mathcal{L}_{s}(\theta) -  \frac{\beta}{M}\sum_{m=N+1}^{N+M}\log(D(\mathbf{X}_m, f_{sdm}(\mathbf{X}_m; \theta);\zeta)) \label{eq:segmentation_network_loss}
 \end{align}
On the other hand, given a fixed segmentation network, we simply minimize the binary cross entropy loss induced by Eqn~\eqref{eq:total_loss} to train the discriminator, i.e.,
$\min_\zeta -\mathcal{V}(\theta,\zeta)$, or $\max_\zeta \mathcal{L}_a(\theta,\zeta)$.
To stablize the overall training, we use an annealing strategy based on a time-dependent Gaussian warm-up function to slowly increase the loss weight $\beta$ (See Sec.~\ref{sec:exp} for details). 
%During training, we update parameters $\theta$ for SN and $\zeta$ for DN alternately. In the early stage, SN generates bad outputs on labeled and unlabled data, thus DN can't improve and evenly destroys the training process. To eliminate the impact of adversarial learning in early stage, we assign $\beta$ as a time-dependent gaussian warming up function. When SN is well trained and able to capture shape representations of labeled data, DN will force SN to be aware of geometric representations of unlabled data. After training several turns, SN generates pretty segmentation and shape representations for both labeled and unlabeled data.
% denotes the dice loss for measuring the segmentation output quality, and $\mathcal{L}_{sdm}$ the mean square error loss to evaluate regression quality of SDM output. 

%% file: section/experiment.tex
\section{Experiments and Results}\label{sec:exp}     

    % table 1
    \begin{table}[t]
        \caption{Quantitative comparisons of semi-supervised segmentation models on the LA dataset. All models use the V-Net as backbone network. Results on two different data partition settings show that our SASSNet outperforms the state-of-the-art results consistently.}
        \label{tab:la_16}
        \centering
        \resizebox{.95\textwidth}{!}{
        \begin{tabular}{c|c|c|c|c|c|c} \hline \hline
        \multirow{2}{*}{Method} & \multicolumn{2}{c|}{\# \textbf{scans used}} & \multicolumn{4}{c}{\textbf{Metrics}}  \\ \cline{2-7}
                                & Labeled        & Unlabeled        & Dice{[}\%{]} & Jaccard{[}\%{]} & ASD{[}voxel{]} & 95HD{[}voxel{]} \\ \hline
        V-Net                   & 80             & 0                & 91.14        & 83.82           & 1.52           & 5.75            \\ \hline  \hline
        V-Net                   & 16             & 0                & 86.03        & 76.06           & 3.51           & 14.26    \\ \hline
        %DAN \cite{zhang2017deep}                     & 16             & 64               & 87.52        & 78.29           & 2.42           & 9.01            \\ \hline
        DAP \cite{zheng2019semi}    & 16    & 64    &87.89 	&78.72  &2.74   &9.29  \\ \hline
        ASDNet \cite{nie2018asdnet}                  & 16             & 64               & 87.90        & 78.85           & \textbf{2.08}           & 9.24            \\ \hline
        TCSE \cite{li2018semi}                    & 16             & 64               & 88.15        & 79.20           & 2.44           & 9.57            \\ \hline
        UA-MT \cite{yu2019uncertainty}                   & 16             & 64               & 88.88        & 80.21           & 2.26           & 7.32            \\ \hline
        UA-MT(+NMS)	          & 16             & 64               & 89.11        & 80.62           & 2.21           & \textbf{7.30}            \\ \hline
        SASSNet(ours)          & 16             & 64               & 89.27        & 80.82           & 3.13           & 8.83            \\ \hline
        SASSNet(+NMS)          & 16             & 64               & \textbf{89.54}        & \textbf{81.24}           & {2.20}           & 8.24            \\ \hline  \hline
        V-Net                   & 8                                           & 0                                    & 79.99          & 68.12           & 5.48           & 21.11           \\  \hline
        DAP \cite{zheng2019semi}    & 8    & 72    &81.89 	&71.23 	&3.80 	&15.81         \\ \hline
        UA-MT \cite{yu2019uncertainty}                   & 8                                           & 72                                   & 84.25          & 73.48           & 3.36           & 13.84           \\ \hline
            UA-MT(+NMS)              & 8                                           & 72                                   & 84.57          & 73.96           & 2.90           & 12.51           \\ \hline
            SASSNet(ours)          & 8                                           & 72                                   & 86.81          & 76.92           & 3.94           & 12.54           \\ \hline
            SASSNet(+NMS)          & 8                                           & 72                                   & \textbf{87.32} & \textbf{77.72}  & \textbf{2.55}  & \textbf{9.62}   \\ \hline \hline

        \end{tabular}}
        \end{table}
    % table 3

We validate our method on the Left Atrium (LA) dataset from Atrial Segmentation Challenge\footnote{http://atriaseg2018.cardiacatlas.org/} with detailed comparisons to prior arts. %and NIH pancreas dataset. For Atrial Segmentation Challenge dataset,
The dataset contains 100 3D gadolinium-enhanced MR imaging scans (GE-MRIs) and LA segmentation masks, with an isotropic resolution of $0.625 \times 0.625 \times 0.625 mm^3$.
Following \cite{yu2019uncertainty}, we split them into 80 scans for training and 20 scans for validation, and apply the same pre-processing methods. 
%For NIH pancreas dataset, there are 82 contrast-enhanced abdominal CT scans as well as corresponding annotations, with varying pixel sizes and slice thickness between 1.5 -- 2.5 mm. We crop the CT scans based on region statistics of interest, clip values to [-100, 240] and normalize them.  Four-fold cross-validation is used to assess the robustness of the model.
%\newline
\subsubsection{Implementation Details and Metrics.} %We train 6000 iterations and evaluate performances on test set. 
The segmentation network is trained by a SGD optimizer for 6000 iterations, with an initial learning rate (lr) 0.01 decayed by 0.1 every 2500 iterations. The discriminator uses $4\times4\times4$ kernels with stride 2 in its convolutional layers and an Adam optimizer with a constant lr 0.0001. %For left atrium dataset, the volume shape is $112 \times 112 \times 80$,
We use a batch size of 4 images and a single GPU with 12Gb RAM for the model training. 
In all our experiments, we set $\alpha$ as 0.3 and $\beta$ as a time-dependent Gaussian warming-up function 
$\lambda (t)=0.001*e^{-5(1-\frac {t} {t_{max}})^2}$ where $t$ indicates number of iterations. %(the maximum value will be a little different according to experiment).

During testing, we take the segmentation map output $\mathbf{M}$ for evaluation. In addition, an non-maximum suppression (NMS)  is applied as the post process in order to remove isolated extraneous regions. We use the standard evaluation metrics, including Dice coefficient (Dice), Jaccard Index (Jaccard), 95\% Hausdorff Distance (95HD) and Average Symmetric Surface Distance (ASD). %\newline

\begin{figure}[t]
	\centering
	\subfigure[2D comparison]{            
		\begin{minipage}[t]{\linewidth}
			\centering
			\includegraphics[width=0.95\linewidth]{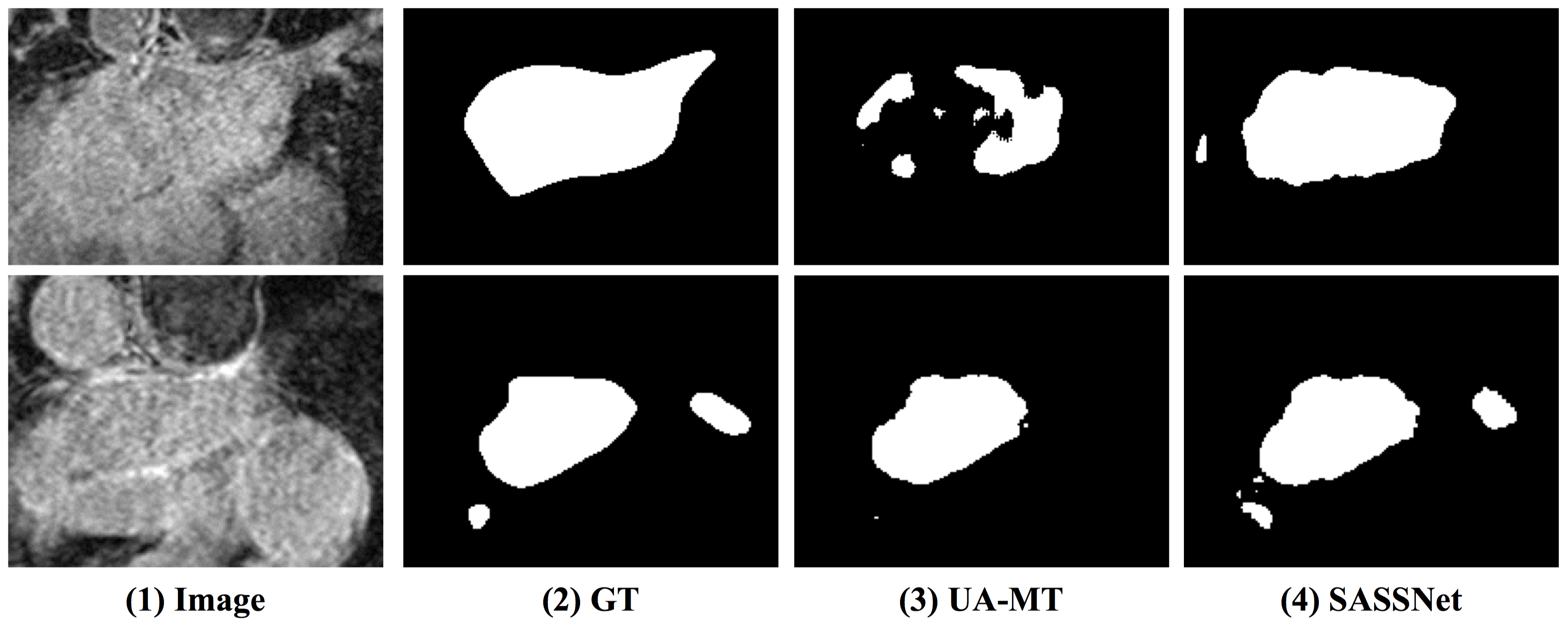}
			%\caption{fig1}
		\end{minipage}%
	}%
	\quad
	\subfigure[3D comparison]{            
		\begin{minipage}[t]{\linewidth}
			\centering
			\includegraphics[width=0.95\linewidth]{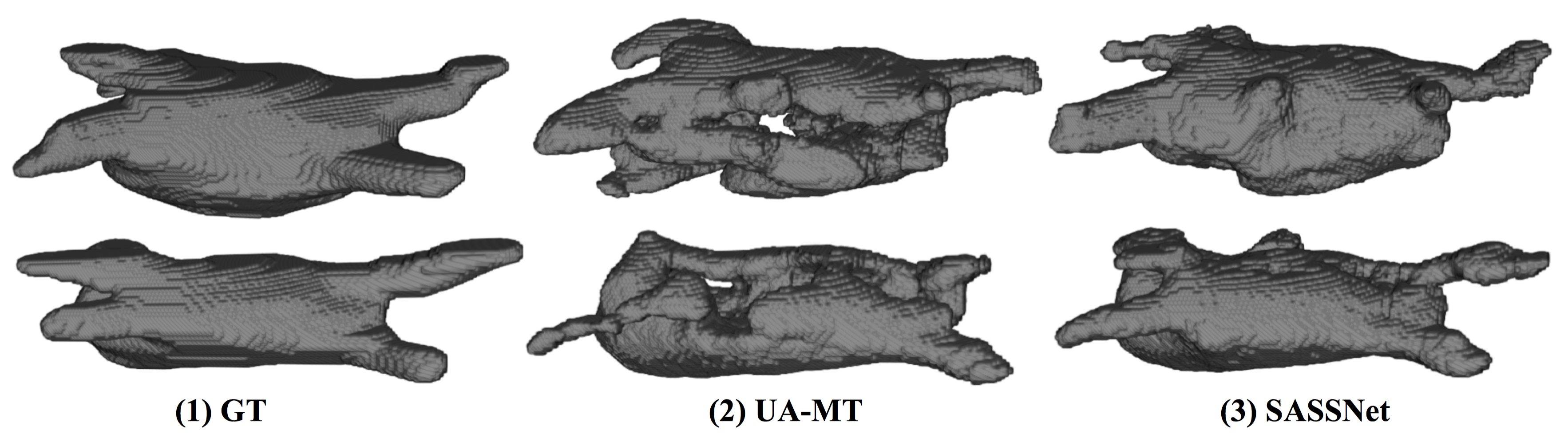}
			%\caption{fig1}
		\end{minipage}%
	}
	\caption{2D and 3D Visualization of the segmentations by UA-MT \cite{yu2019uncertainty} and our method, where GT denotes groundtruth segmetnation. %Our SASSNet generates complete segmentation with good global shape, surpass UA-MT by a large margin.
	}
	\label{fig:la_vis2d}
\end{figure}

% \subsection{Experiments on LA dataset}
\subsubsection{Quantitative Evaluation and Comparison.} 

We evaluate our method in two different settings with comparisons to several recent semi-supervised segmentation approaches, including DAP \cite{zheng2019semi}, ASDNet \cite{nie2018asdnet}, TCSE \cite{li2018semi} and UA-MT \cite{yu2019uncertainty}. Table \ref{tab:la_16} presents a summary of the quantitative results, in which we first show the upper-bound performance achieved by a fully-supervised network, followed by two individual settings. 

The first setting follows the work \cite{yu2019uncertainty}, which takes 20\% of training data as labeled data (16 labeled), and the others as unlabeled data for semi-supervised training. We can see that this setting is relative easy as the model trained with 20\% of data already achieves good performance (86.03\% in Dice). Among the semi-supervised methods, the DAP performs worst, indicating the limitation of an atlas-based prior, while UA-MT achieves the top performance in the previous methods. Our method outperforms all the other semi-supervised networks in both Dice (89.54\%) and Jaccard (81.24\%), and achieves competitive results on other metrics. In particular, our SASSNet surpasses UA-MT in Dice without resorting to a complex multiple network architecture.     
  
To validate the robustness of our method, we also consider a more challenging setting in which we only have 8 labeled images for training. The second half of Table \ref{tab:la_16} show the comparison results, where SASSNet outperforms UA-MT with a large margin (Dice: +2.56\% without NMS and +3.07\% with NMS). Without NMS, our SASSNet tends to generate more foreground regions, which leads to slightly worse performance on ASD and 95HD. However, it also produce better segmentation preserving the original object shape. By contrast, UA-MT often misses inner regions of target objects and generates irregular shapes. Figure~\ref{fig:la_vis2d} provides several qualitative results for visual comparison.

%To further validate the improvement of SASSNet, we continually decrease the labeled number to 8. In Table \ref{tab:la_16}, SASSNet outperform UA-MT with a large margin (Dice: +2.56\% without NMS and +3.07\% with NMS). In the absence of NMS, SASSNet presents inferior performance on ASD and 95HD, because it tend to generate extraneous foregrounds when preserving the original overall shape. On the other hand, UA-MT usually miss inner foreground regions and care less for the global shape. The visualization comparison is illustrate in Fig. \ref{fig:la_vis2d}. We find that SASSNet perform well in capturing the global shape, as well as generating smooth and complete segmentations, whereas UA-MT concerns only local regions thus produces inferior output. In short, our SASSNet are proved to be powerful in capturing global geometry shape, and reports state-of-the-art performances in semi-supervised segmentation. %\newline

 \begin{table}[t]
	\caption{Effectiveness of our proposed modules on the LA dataset. All the models use the same V-Net as the backbone, and we conduct an ablative study to show the contribution of each component module.}
	\label{tab:la_ablation}
	\resizebox{\textwidth}{!}{
		\begin{tabular}{l|c|c|c|c|c|c|c} \hline \hline
			\multirow{2}{*}{Method} & \multicolumn{2}{c|}{\# \textbf{scans used}} & \multicolumn{4}{c|}{\textbf{Metrics}} & \multicolumn{1}{c}{\textbf{Cost}} \\ \cline{2-8}
			& Labeled        & Unlabeled        & Dice{[}\%{]} & Jaccard{[}\%{]} & ASD{[}voxel{]} & 95HD{[}voxel{]} & Params{[}M{]} \\ \hline
			V-Net           & 8 & 0  & 79.99 & 68.12 & 5.48 & 21.11 & 187.7  \\ \hline
			V-Net +SDM      & 8 & 0 & 81.12 & 69.75 & 6.93 & 25.58 & 187.9 \\ \hline 
			V-Net +SDM +GAN & 8 & 72 & 86.81 & 76.92 & 3.94 & 12.54 & 249.7 \\ \hline 
			UA-MT \cite{yu2019uncertainty} & 8 & 72 & 84.25  & 73.48  & 3.36 & 13.84 & 375.5 \\ \hline
			V-Net +SDM +MT  & 8 & 72 & 84.97 & 74.14 & 6.12 & 22.20 & 375.8 \\ \hline
			\hline
		\end{tabular}
	}
\end{table}

  % figure: vis_la

\subsubsection{Ablative Study.} We conduct several detailed experimental studies to examine the effectiveness of our proposed SDM head and the adversarial loss (GAN). Table \ref{tab:la_ablation} shows the quantitative results of different model settings. 
The first row is a V-Net trained with only the labeled data, which is our base model. We first add a SDM head, denoted as V-Net+SDM, and as shown in the second row, such joint learning improves segmentation results by 1.1\% in Dice. We then add the unlabeled data and our adversarial loss, denoted as V-Net+SDM+GAN, which significantly improves the performance (5.7\% in Dice). 

We also compare our semi-supervised learning strategy with two methods in the mean-teacher (MT) framework (last two rows). One is the original UA-MT and the other is our segmentation network with the MT consistency loss. Our SASSNet outperforms both methods with higher Dice and Jaccard scores, which indicates the advantage of our representation and loss design. Moreover, our network has a much simpler architecture than those two networks.

%SASSNet outperform UA-MT on all metrics with less parameters, showing that heavy mean-teacher framework is not necessary and adversarial learning apporoach a better choice. %\newline

%Furthermore, a GAN is added for the last row. V-Net with SDM head and consistencies brings 4.98\% Dice improvement, and GAN gains a further 1.84\% Dice. It means that SDM head equips the segmentation model with global shape awareness, and GAN enforces geometric awareness for unlabeled data. Additionally, unlabeled volumes and GAN prevent the model from overfitting in the scenario of limited labeled data.
    
%\subsection{Experiment on NIH pancreas}
%Furthermore, We conduct experiments on NIH pancreas dataset, which is more challenging because of their large variations in shape and size. Fully supervised state-of-the-art approach \cite{zhao_AutoPancreas_2019} achieves a mean Dice 85.99\%, while our concern is the effectiveness of SASSNet in semi-supervised learning. Without other tricks, we use V-Net as the baseline model, and Table \ref{tab:pan_exp} shows the quantitative improvements when applying SASSNet.  The upper bound is illustrated in the first row. When continually decreasing the labeled data, SASSNet always outperform V-Net by a significant gap in all metrics, demonstrating the generalization ability of utilizing unlabeled data and geometric awareness. Finally, Fig. \ref{fig:pan_vis3d} visualizes the our improvements on 3D global shape.

%% file: section/conclusion.tex
\section{Conclusion}
In this paper, we proposed a shape-aware semi-supervised segmentation approach for 3D medical scans. In contrast to previous methods, our method exploits the regularity in geometric shapes of the target object class for effective segment prediction and network learning. We developed a multi-task segmentation network that jointly predicts semantic segmentation and SDM of object surfaces, and a semi-supervised learning loss enforcing consistency between the predicted SDMs of labeled and unlabeled data. We validated our approach on the Atrial Segmentation Challenge dataset, which demonstrates that our segmentation network outperforms the state-of-the-art methods and generates object segmentation with high-quality global shapes.